\documentclass[a4,a4wide]{article} % For LaTeX2e

\usepackage{aaai}
\usepackage{amsfonts}
\usepackage{amsmath} % assumes amsmath package installed
\usepackage{amssymb}  % assumes amsmath package installed
\usepackage{graphicx}
\usepackage{graphics} % for pdf, bitmapped graphics files
\usepackage{mathptmx} % assumes new font selection scheme installed
\usepackage{url}
\usepackage{rotating}
\usepackage{subfigure}
\usepackage{wrapfig}
\usepackage{multicol}
\usepackage[linesnumbered,boxruled,lined]{algorithm2e}

\def\no{\; {not} \;}

\newcommand{\stt}[1]{{\small\texttt{#1}}}

\newtheorem{example}{Example}
\newtheorem{definition}{Definition}
\newtheorem{proposition}{Proposition}
\newcommand{\rif}{\stackrel{\,\,+}{\leftarrow}}
\newenvironment{small_ind_s_itemize}{\begin{list}{$\bullet$}
{\setlength{\rightmargin}{0em}
\setlength{\leftmargin}{1em}
\setlength{\itemsep}{0em}
\setlength{\topsep}{0em}
\setlength{\parsep}{0em}}}{\end{list}}

\newcounter{ctr}
\newenvironment{s_enumerate}{\begin{list}{\thectr.}
{\usecounter{ctr}
\setlength{\rightmargin}{0.3cm}
\setlength{\leftmargin}{0.3cm}
\setlength{\itemsep}{0em}
\setlength{\topsep}{0em}
\setlength{\itemindent}{0.5cm}
\setlength{\parsep}{0em}}}{\end{list}}

\title{KR$^3$: An Architecture for Knowledge Representation and
  Reasoning in Robotics}

\author{
Shiqi Zhang \\
Department of Computer Science\\
Texas Tech University, USA\\
\texttt{shiqi.zhang6@gmail.com} \\
\And
Mohan Sridharan \\
Department of Computer Science\\
Texas Tech University, USA\\
\texttt{mohan.sridharan@ttu.edu} \\
\AND
Michael Gelfond \\
Department of Computer Science\\
Texas Tech University, USA\\
\texttt{michael.gelfond@ttu.edu} 
\And
Jeremy Wyatt \\
School of Computer Science\\
University of Birmingham, UK\\
\texttt{jlw@cs.bham.ac.uk} 
}

\begin{document}

\nocopyright

\maketitle

%%%%%%%%%%%%%%%%%%%%%%%%%%%%%%%%%%%%%%%%%%%%%%%%%%%%%%%%%%%%%%%%%%%%%%%%%%%%%%
%%%%%%%%%%%%%%%%%%%%%%%%%%%%%%%%%%%%%%%%%%%%%%%%%%%%%%%%%%%%%%%%%%%%%%%%%%%%%%

\begin{abstract}
  This paper describes an architecture that combines the complementary
  strengths of declarative programming and probabilistic graphical
  models to enable robots to represent, reason with, and learn from,
  qualitative and quantitative descriptions of uncertainty and
  knowledge. An action language is used for the low-level (LL) and
  high-level (HL) system descriptions in the architecture, and the
  definition of recorded histories in the HL is expanded to allow
  prioritized defaults. For any given goal, tentative plans created in
  the HL using default knowledge and commonsense reasoning are
  implemented in the LL using probabilistic algorithms, with the
  corresponding observations used to update the HL history. Tight
  coupling between the two levels enables automatic selection of
  relevant variables and generation of suitable action policies in the
  LL for each HL action, and supports reasoning with violation of
  defaults, noisy observations and unreliable actions in large and
  complex domains.  The architecture is evaluated in simulation and on
  physical robots transporting objects in indoor domains; the benefit
  on robots is a reduction in task execution time of $39\%$ compared
  with a purely probabilistic, but still hierarchical, approach.
\end{abstract}

%%%%%%%%%%%%%%%%%%%%%%%%%%%%%%%%%%%%%%%%%%%%%%%%%%%%%%%%%%%%%%%%%%%%%%%%%%%%%%
%%%%%%%%%%%%%%%%%%%%%%%%%%%%%%%%%%%%%%%%%%%%%%%%%%%%%%%%%%%%%%%%%%%%%%%%%%%%%%
\section{Introduction}
\label{sec:intro}
Mobile robots deployed in complex domains receive far more raw data
from sensors than is possible to process in real-time, and may have
incomplete domain knowledge. Furthermore, the descriptions of
knowledge and uncertainty obtained from different sources may
complement or contradict each other, and may have different degrees of
relevance to current or future tasks.  Widespread use of robots thus
poses fundamental knowledge representation and reasoning
challenges---robots need to represent, learn from, and reason with,
qualitative and quantitative descriptions of knowledge and
uncertainty. Towards this objective, our architecture combines the
knowledge representation and non-monotonic logical reasoning
capabilities of declarative programming with the uncertainty modeling
capabilities of probabilistic graphical models.  The architecture
consists of two tightly coupled levels and has the following key
features:
\begin{s_enumerate}
\item An action language is used for the HL and LL system descriptions
  and the definition of recorded history is expanded in the HL to
  allow prioritized defaults.
\item For any assigned objective, tentative plans are created in the
  HL using default knowledge and commonsense reasoning, and
  implemented in the LL using probabilistic algorithms, with the
  corresponding observations adding suitable statements to the HL
  history.
\item For each HL action, abstraction and tight coupling between the
  LL and HL system descriptions enables automatic selection of
  relevant variables and generation of a suitable action policy in the
  LL.
\end{s_enumerate}
In this paper, the HL domain representation is translated into an
Answer Set Prolog (ASP) program, while the LL domain representation is
translated into partially observable Markov decision processes
(POMDPs). The novel contributions of the architecture, e.g., allowing
histories with prioritized defaults, tight coupling between the two
levels, and the resultant automatic selection of the relevant
variables in the LL, support reasoning with violation of defaults,
noisy observations and unreliable actions in large and complex
domains. The architecture is grounded and evaluated in simulation and
on physical robots moving objects in indoor domains.

%%%%%-----------------------------------------------------------------------------
%%%%%-----------------------------------------------------------------------------
\section{Related Work}
\label{sec:relwork}
Probabilistic graphical models such as POMDPs have been used to
represent knowledge and plan sensing, navigation and interaction for
robots~\cite{Hoey:CVIU10,Rosenthal:aaai12}. However, these
formulations (by themselves) make it difficult to perform commonsense
reasoning, e.g., default reasoning and non-monotonic logical
reasoning, especially with information not directly relevant to tasks
at hand. In parallel, research in classical planning has provided many
algorithms for knowledge representation and logical
reasoning~\cite{Ghallab:plan04}, but these algorithms require
substantial prior knowledge about the domain, task and the set of
actions. Many of these algorithms also do not support merging of new,
unreliable information from sensors and humans with the current
beliefs in a knowledge base.  Answer Set Programming (ASP), a
non-monotonic logic programming paradigm, is well-suited for
representing and reasoning with commonsense
knowledge~\cite{Gelfond:book08,Baral:book03}. An international
research community has been built around ASP, with applications such
as reasoning in simulated robot housekeepers and for representing
knowledge extracted from natural language human-robot
interaction~\cite{Chen:HRI12,Erdem:ISR12}.  However, ASP does not
support probabilistic analysis, whereas a lot of information available
to robots is represented probabilistically to quantitatively model the
uncertainty in sensor input processing and actuation in the real
world.

Researchers have designed cognitive
architectures~\cite{Laird:AI87,Langley:aaai06,Talamadupula:TIST10},
and developed algorithms that combine deterministic and probabilistic
algorithms for task and motion planning on
robots~\cite{Kaelbling:IJRR13,Hanheide:ijcai11}. Recent work has also
integrated ASP and POMDPs for non-monotonic logical inference and
probabilistic planning on robots~\cite{zhang:icdl12}. Some examples of
principled algorithms developed to combine logical and probabilistic
reasoning include probabilistic first-order
logic~\cite{halpern:book03}, first-order relational
POMDPs~\cite{Sanner:aaai10}, Markov logic
network~\cite{Richardson:ML06}, Bayesian
logic~\cite{milch:bookchap07}, and a probabilistic extension to
ASP~\cite{baral:TPLP09}. However, algorithms based on first-order
logic for probabilistically modeling uncertainty do not provide the
desired expressiveness for capabilities such as default reasoning,
e.g., it is not always possible to express uncertainty and degrees of
belief quantitatively.  Other algorithms based on logic programming
that support probabilistic reasoning do not support one or more of the
desired capabilities: reasoning as in causal Bayesian networks;
incremental addition of probabilistic information; reasoning with
large probabilistic components; and dynamic addition of variables with
different ranges~\cite{baral:TPLP09}. The architecture described in
this paper is a step towards achieving these capabilities. It exploits
the complementary strengths of declarative programming and
probabilistic graphical models to represent, reason with, and learn
from qualitative and quantitative descriptions of knowledge and
uncertainty, enabling robots to automatically plan sensing and
actuation in larger domains than was possible before.

%%%%%-----------------------------------------------------------------------------
%%%%%-----------------------------------------------------------------------------
\section{KRR Architecture}
\label{sec:arch}
This section describes our architecture's HL and LL domain
representations. The syntax, semantics and representation of the
corresponding transition diagrams are described in an \emph{action
  language} AL~\cite{Gelfond:aibook14}.  Action languages are formal
models of parts of natural language used for describing transition
diagrams. AL has a sorted signature containing three \emph{sorts}:
\stt{statics}, \stt{fluents} and \stt{actions}.  Statics are domain
properties whose truth values cannot be changed by actions, while
fluents are properties whose truth values are changed by actions.
Actions are defined as a set of elementary actions that can be
executed in parallel. A domain property \stt{p} or its negation $\lnot
p$ is a domain literal.  AL allows three types of statements:
\begin{align}
  \begin{array}{ll}
    a~~\mathbf{causes}~~l_{in}~~\mathbf{if}~p_0,\ldots,p_m\qquad \qquad
\qquad~~~~\textrm{(Causal law)} \\ \nonumber 
    l~~\mathbf{if}~~p_0,\ldots,p_m\qquad \qquad \qquad \qquad ~~~~~~~\textrm{(State constraint)}\\\nonumber 
    \mathbf{impossible}~~a_0,\ldots,a_k~~\mathbf{if}~~p_0,\ldots,p_m\qquad 
    \\\nonumber \qquad \qquad \qquad \qquad \qquad \qquad~~~\textrm{(Executability condition)}
  \end{array}
\end{align}
where $a$ is an action, $l$ is a literal, $l_{in}$ is a inertial
fluent literal, and $p_0,\ldots,p_m$ are domain literals. The causal
law states, for instance, that action $a$ causes inertial fluent
literal $l_{in}$ if the literals $p_0,\ldots,p_m$ hold true. A
collection of statements of AL forms a system/domain description.

As an illustrative example used throughout this paper, we will
consider a robot that has to move objects to specific places in an
indoor domain. The domain contains four specific places:
\emph{office}, \emph{main\_library}, \emph{aux\_library}, and
\emph{kitchen}, and a number of specific objects of the sorts:
\emph{textbook}, \emph{printer} and \emph{kitchenware}.

%%%%%-----------------------------------------------------------------------------
\subsection{HL domain representation}
\label{sec:arch-hl}
\begin{figure}[tbc]
  \begin{center}
    \includegraphics[width=1\columnwidth]{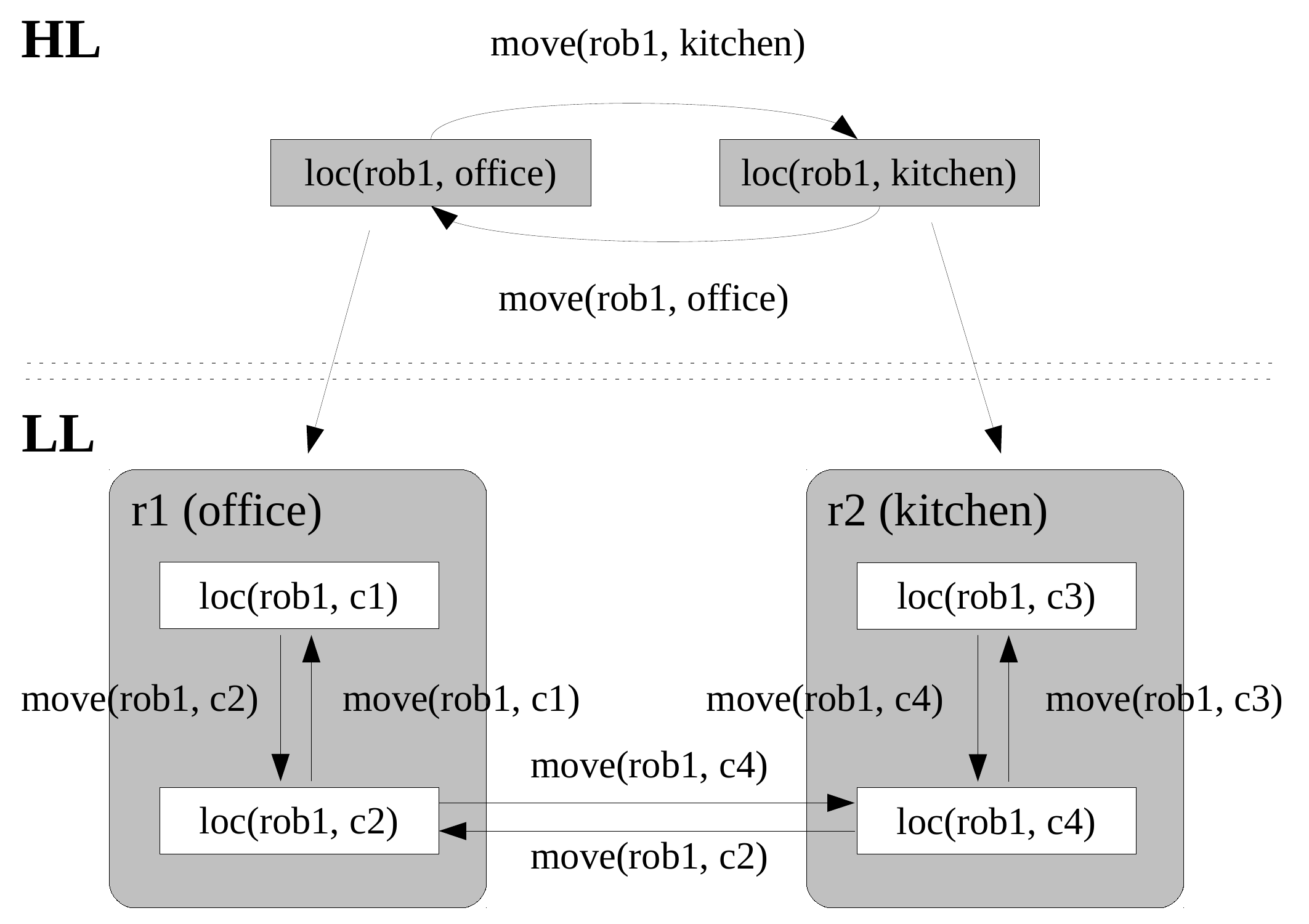}
    \caption{Illustrative example of state transitions in the HL and
      the LL of the architecture.}
  \label{fig:state-transition}
  \end{center}
\end{figure}

The HL domain representation consists of a system description
$\mathcal{D}_H$ and histories with defaults $\mathcal{H}$.
$\mathcal{D}_H$ consists of a sorted signature and axioms used to
describe the HL transition diagram $\tau_H$. The sorted signature:
$\Sigma_H=\langle\mathcal{O}, \mathcal{F}, \mathcal{P}\rangle$ is a
tuple that defines the names of objects, functions, and predicates
available for use in the HL. The sorts in our example are:
\stt{place}, \stt{thing}, \stt{robot}, and \stt{object}; \stt{object}
and \stt{robot} are subsorts of \stt{thing}. Robots can move on their
own, but objects cannot move on their own. The sort \stt{object} has
subsorts such as \stt{textbook}, \stt{printer} and \stt{kitchenware}.
The fluents of the domain are defined in terms of their arguments:
\begin{align}
  &loc(thing, place)\\ \nonumber
  &in\_hand(robot, object)%\\ \nonumber
%  &same\_loc(thing, thing)
\end{align}
The first predicate states the location of a thing; and the second
predicate states that a robot has an object.% ; and the third predicate
% states that two things have the same location. 
These two predicates are \emph{inertial fluents} subject to the law of
inertia, which can be changed by an action. The \emph{actions} in this
domain include:
\begin{align}
  &move(robot, place)\\ \nonumber
  &grasp(robot, object)\\ \nonumber
  &putdown(robot, object)
\end{align}
The dynamics of the domain are defined using the following causal
laws:
\begin{align}
  &move(robot, Pl)~~\mathbf{causes}~~loc(robot, Pl) \\ \nonumber
  &grasp(robot, Ob)~~\mathbf{causes}~~in\_hand(robot, Ob)\\\nonumber %~~\mathbf{if}~~loc(robot, Pl), loc(Ob, Pl)\\\nonumber
  &putdown(robot, Ob)~~\mathbf{causes}~~\neg in\_hand(robot, Ob)
\end{align}
state constraints:
\begin{align}
  &loc(Ob, Pl)~~\mathbf{if}~~loc(robot, Pl),~~in\_hand(robot, Ob) \\\nonumber
  &\neg loc(Th, Pl_1)~~\mathbf{if}~~loc(Th, Pl_2),~Pl_1\neq Pl_2 %\\\nonumber
%  &same\_loc(Th1, Th2)~~\mathbf{if}~~loc(Th1, Pl),~loc(Th2, Pl)
\end{align}
and executability conditions:
\begin{align}
  &\mathbf{impossible}~~move(robot, Pl)~~\mathbf{if}~~loc(robot, Pl) \\\nonumber
  &\mathbf{impossible}~~A_1,~A_2,~~\mathbf{if}~~A_1\neq A_2. \\\nonumber
%  &\mathbf{impossible}~~grasp(robot, Ob)~~\mathbf{if}~~\neg same\_loc(robot, Ob) \\\nonumber
  &\mathbf{impossible}~~grasp(robot, Ob)~~\mathbf{if}~~loc(robot, Pl1), \\\nonumber
  &~~~~~~~~~~~~~~~~~~~~loc(Ob, Pl2), Pl1\neq Pl2  \\\nonumber
  &\mathbf{impossible}~~grasp(robot, Ob)~~\mathbf{if}~~ in\_hand(robot, Ob) \\\nonumber
  &\mathbf{impossible}~~putdown(robot, Ob)~~\mathbf{if}~~\neg in\_hand(robot, Ob)
\end{align}
% where the first condition prevents a robot from moving to a location
% if it is already there; the second condition prevents concurrent
% execution of actions; the third condition requires the robot to be in
% the same location as an object before grasping it; the fourth
% condition prevents a robot from grasping an object if it already has
% an object; while the fifth condition requires the robot to hold an
% object before putting it down.
The top part of Figure~\ref{fig:state-transition} shows some state
transitions in the HL; nodes include a subset of fluents (robot's
position) and actions are the arcs between nodes. Although
$\mathcal{D}_H$ does not include the costs of executing actions, these
are included in the LL (see Section~\ref{sec:arch-ll}).

%%%%%-----------------------------------------------------------------------------
\subsubsection{Histories with defaults}
\label{sec:arch-hl-hist}
A recorded history of a dynamic domain is usually defined as a
collection of records of the form $obs(fluent,boolean,step)$ and
$hpd(action,step)$. The former states that a specific fluent was
observed to be true or false at a given step of the domain's
trajectory, and the latter states that a specific action happened (or
was executed by the robot) at that step. In this paper, \emph{we
  expand on this view by allowing histories to contain (possibly
  prioritized) defaults describing the values of fluents in their
  initial states}.  A default $d(X)$ stating that \emph{in the typical
  initial state elements of class $c$ satisfying property $b$ also
  have property $p$} is represented as:
\begin{equation}\label{d}
d(X) = \left\{
\begin{array}{l}
default(d(X))\\
head(d(X),p(X))\\
body(d(X),c(X))\\
body(d(X),b(X))
\end{array}
\right.
\end{equation}
where the literal in the ``head'' of the default, e.g., $p(X)$ is true
if all the literals in the ``body'' of the default, e.g., $b(X)$ and
$c(X)$, hold true; see~\cite{Gelfond:aibook14} for formal semantics of
defaults. In this paper, we abbreviate $obs(f,true,0)$ and
$obs(f,false,0)$ as $init(f,true)$ and $init(f,false)$ respectively.

\begin{example}\label{ex1}[Example of defaults]\\
  {\rm Consider the following statements about the locations of
    textbooks in the initial state in our illustrative example.
    \emph{Textbooks are typically in the main library. If a textbook
      is not there, it is in the auxiliary library.  If a textbook is
      checked out, it can be found in the office.} These defaults can
    be represented as:

\begin{equation}
  \label{d1}
\begin{array}{ll}
default(d_1(X)) \\
head(d_1(X),loc(X,main\_library)) \\
body(d_1(X),textbook(X))
\end{array}
\end{equation}

\begin{equation}\label{d2}
\begin{array}{l}
default(d_2(X))\\
head(d_2(X),loc(X,aux\_library))\\
body(d_2(X),textbook(X))\\
body(d_2(X),\neg loc(X,main\_library))
\end{array}
\end{equation}

\begin{equation}\label{d3}
\begin{array}{l}
default(d_3(X))\\
head(d_3(X),loc(X,office))\\
body(d_3(X),textbook(X))\\
body(d_3(X),\lnot loc(X, main\_library)) \\
body(d_3(X),\lnot loc(X, aux\_library))
\end{array}
\end{equation}

\noindent 
A default such as ``kitchenware are usually in the kitchen'' may be
represented in a similar manner. We first present multiple informal
examples to illustrate reasoning with these defaults;
Definition~\ref{def3} (below) will formalize this reasoning.  For
textbook $tb_1$, history $\mathcal{H}_1$ containing the above
statements should entail: $holds(loc(tb_1,main\_library),0)$. A
history $\mathcal{H}_2$ obtained from $\mathcal{H}_1$ by adding an
observation: $init(loc(tb_1,main\_library),false)$ renders the first
default inapplicable; hence $\mathcal{H}_2$ should entail:
$holds(loc(tb_1, aux\_library),0)$.  A history $\mathcal{H}_3$
obtained from $\mathcal{H}_2$ by adding an observation:
$init(loc(tb_1, aux\_library), false)$ entails:
$holds(loc(tb_1,office),0)$.

Consider history $\mathcal{H}_4$ obtained by adding observation:
$obs(loc(tb_1,main\_library),false,1)$ to $\mathcal{H}_1$. This
observation should defeat the default $d_1$ in Equation~\ref{d1}
because if this default's conclusion were true in the initial state,
it would also be true at step $1$ (by inertia), which contradicts our
observation. The book $tb_1$ is thus not in the main library
initially. The second default will conclude that this book is
initially in the auxiliary library---the inertia axiom will propagate
this information and $\mathcal{H}_4$ will entail: $holds(loc(tb_1,
aux\_library),1)$.  }
\end{example}
The definition of entailment relation can now be given with respect to
a fixed system description $\mathcal{D}_H$.  We start with the notion
of \emph{a state of transition diagram $\tau_H$ of $\mathcal{D}_H$
  compatible with a description $\mathcal{I}$ of the initial state of
  history $\mathcal{H}$}. We use the following terminology. We say
that a set $S$ of literals is \emph{closed under a default} $d$ if $S$
contains the head of $d$ whenever it contains all literals from the
body of $d$ and does not contain the literal contrary to $d$'s head.
$S$ is \emph{closed under a constraint} of $\mathcal{D}_H$ if $S$
contains the constraint's head whenever it contains all literals from
the constraint's body. Finally, we say that a set $U$ of literals is
the \emph{closure of $S$} if $S \subseteq U$, $U$ is closed under
constraints of $\mathcal{D}_H$ and defaults of $\mathcal{H}$, and no
proper subset of $U$ satisfies these properties.

\begin{definition}\label{def1}[Compatible initial states]\\ 
  \rm{A state $\sigma$ of $\tau_H$ is \emph{compatible} with a
    description $\mathcal{I}$ of the initial state of history
    $\mathcal{H}$ if:
\begin{s_enumerate}
\item $\sigma$ satisfies all observations of $\mathcal{I}$,
\item $\sigma$ contains the closure of the union of statics of
  $\mathcal{D}_H$ and the set $\{f : init(f,true) \in \mathcal{I}\} \cup
  \{\neg f : init(f,false) \in \mathcal{I}\}$.
\end{s_enumerate}
Let $\mathcal{I}_k$ be the description of the initial state of history
$\mathcal{H}_k$. States in Example~\ref{ex1} compatible with
$\mathcal{I}_1$, $\mathcal{I}_2$, $\mathcal{I}_3$ must then contain
$\{loc(tb_1,main\_library)\}$, $\{loc(tb_1,aux\_library)\}$, and
$\{loc(tb_1,office)\}$ respectively.  There are multiple such states,
which differ by the location of robot.  Since $\mathcal{I}_1 =
\mathcal{I}_4$ they have the same compatible states.  Next, we define
\emph{models} of history $\mathcal{H}$, i.e., paths of the transition
diagram $\tau_H$ of $\mathcal{D}_H$ compatible with $\mathcal{H}$.  }
\end{definition}

\begin{definition}\label{def2}[Models]\\ 
  \rm{A path $P$ of $\tau_H$ is a \emph{model} of history
    $\mathcal{H}$ with description $\mathcal{I}$ of its initial state
    if there is a collection $E$ of $init$ statements such that:
\begin{s_enumerate}
\item If $init(f,true) \in E$ then $\neg f$ is the head of one of the
  defaults of $\mathcal{I}$. Similarly, for $init(f,false)$.
\item The initial state of $P$ is compatible with the description:
  $\mathcal{I}_E = \mathcal{I} \cup E$.
\item Path $P$ satisfies all observations in $\mathcal{H}$.
\item There is no collection $E_0$ of $init$ statements which has less
  elements than $E$ and satisfies the conditions above.
\end{s_enumerate}
We will refer to $E$ as an \emph{explanation} of $\mathcal{H}$. Models
of $\mathcal{H}_1$, $\mathcal{H}_2$, and $\mathcal{H}_3$ are paths
consisting of initial states compatible with $\mathcal{I}_1$,
$\mathcal{I}_2$, and $\mathcal{I}_3$---the corresponding explanations
are empty. However, in the case of $\mathcal{H}_4$, the situation is
different---the predicted location of $tb_1$ will be different from
the observed one.  The only explanation of this discrepancy is that
$tb_1$ is an exception to the first default. Adding $E =
\{init(loc(tb_1,main\_library),false)\}$ to $\mathcal{I}_4$ will
resolve the problem. }
\end{definition}

\begin{definition}\label{def3}[Entailment and consistency]
  \rm{
\begin{small_ind_s_itemize}
\item Let $\mathcal{H}^n$ be a history of length $n$, $f$ be a fluent,
  and $0\leq i \leq n$ be a step of $\mathcal{H}^n$. We say that
  $\mathcal{H}^n$ \emph{entails} a statement $Q = holds(f,i)$ ($\neg
  holds(f,i)$) if for every model $P$ of $\mathcal{H}^n$, fluent
  literal $f$ ($\neg f$) belongs to the $i$th state of $P$. We denote
  the entailment as $\mathcal{H}^n \models Q$.
\item A history which has a model is said to be \emph{consistent}.
\end{small_ind_s_itemize}
}
\end{definition}
It can be shown that histories from Example~\ref{ex1} are consistent
and that our entailment captures the corresponding intuition.

%%%%%-----------------------------------------------------------------------------
\subsubsection{Reasoning with HL domain representation}
\label{sec:arch-hl-entail}
The HL domain representation ($\mathcal{D}_H$ and $\mathcal{H}$) is
translated into a program in CR-Prolog, which incorporates consistency
restoring rules in ASP~\cite{Balduccini:aaaisymp03,Gelfond:aibook14};
specifically, we use the knowledge representation language SPARC that
expands CR-Prolog and provides explicit constructs to specify objects,
relations, and their sorts~\cite{Balai:lpnmr13}. ASP is a declarative
language that can represent recursive definitions, defaults, causal
relations, special forms of self-reference, and other language
constructs that occur frequently in non-mathematical domains, and are
difficult to express in classical logic
formalisms~\cite{Baral:book03}. ASP is based on the stable model
semantics of logic programs, and builds on research in non-monotonic
logics~\cite{Gelfond:book08}. A CR-Prolog program is thus a collection
of statements describing domain objects and relations between them.
The ground literals in an \emph{answer set} obtained by solving the
program represent beliefs of an agent associated with the
program\footnote{SPARC uses DLV~\cite{Leone:TOCL06} to generate answer
  sets.}; program consequences are statements that are true in all
such belief sets.  Algorithms for computing the entailment relation of
AL and related tasks such as planning and diagnostics are thus based
on reducing these tasks to computing answer sets of programs in
CR-Prolog. First, $\mathcal{D}_H$ and $\mathcal{H}$ are translated
into an ASP program $\Pi(\mathcal{D}_H,\mathcal{H})$ consisting of
direct translation of causal laws of $\mathcal{D}_H$, inertia axioms,
closed world assumption for defined fluents, reality checks, records
of observations, actions and defaults from $\mathcal{H}$, and special
axioms for $init$: 
\begin{align}
&holds(F,0) \leftarrow init(F,true)\\ \nonumber
&\neg holds(F,0) \leftarrow init(F,false)
\end{align}
In addition, every default of $\mathcal{I}$ is turned into an ASP
rule:
\begin{align}
  holds(p(X),0) \leftarrow &c(X),~~holds(b(X),0),\\\nonumber
  &\no \neg holds(p(X),0)
\end{align}
and a consistency-restoring rule:
\begin{align}
  \neg holds(p(X),0) \rif & c(X),~~holds(b(X),0)
\end{align}
which states that to restore consistency of the program one may assume
that the conclusion of the default is false. For more details about
the translation, CR-rules and CR-Prolog, please
see~\cite{Gelfond:aibook14}.

\begin{proposition}\label{prop1}[Models and Answer Sets]\\ 
  A path $P = \langle
  \sigma_0,a_0,\sigma_1,\dots,\sigma_{n-1},a_n\rangle$ of $\tau_H$ is
  a model of history $\mathcal{H}^n$ iff there is an answer set $S$ of
  a program $\Pi(\mathcal{D}_H,\mathcal{H})$ such that:
\begin{s_enumerate}
\item A fluent $f \in \sigma_i$ iff $holds(f,i) \in S$,
\item A fluent literal $\neg f \in \sigma_i$ iff $\neg holds(f,i) \in
  S$,
\item An action $e \in a_i$ iff $occurs(e,i) \in S$.
\end{s_enumerate}
\end{proposition}
The proposition reduces computation of models of $\mathcal{H}$ to
computing answer sets of a CR-Prolog program. This proposition allows
us to reduce the task of planning to computing answer sets of a
program obtained from $\Pi(\mathcal{D}_H,\mathcal{H})$ by adding the
definition of a goal, a constraint stating that the goal must be
achieved, and a rule generating possible future actions of the robot.

%%%%%-----------------------------------------------------------------------------
\subsection{LL domain representation}
\label{sec:arch-ll}
The LL system description $\mathcal{D}_L$ consists of a sorted
signature and axioms that describe a transition diagram $\tau_L$.  The
sorted signature $\Sigma_L$ of action theory describing $\tau_L$
includes the sorts from signature $\Sigma_H$ of HL with two additional
sorts \stt{room} and \stt{cell}, which are subsorts of sort
\stt{place}. Their elements satisfy the static relation
\emph{part\_of(cell, room)}. We also introduce the static
\emph{neighbor(cell, cell)} to describe neighborhood relation between
cells. Fluents of $\Sigma_L$ include those of $\Sigma_H$, an
additional inertial fluent: \emph{searched(cell, object)}---robot
searched a cell for an object---and two defined fluents:
\emph{found(object, place)}---an object was found in a place---and
\emph{continue\_search(room, object)}---the search for an object is
continued in a room.
% Objects in the LL are also characterized by features (i.e.,
% properties) extracted from images, e.g., color, shape, and local
% gradients.

The actions of $\Sigma_L$ include the HL actions that are viewed as
being represented at a higher resolution, e.g., movement is possible
to specific cells. The causal law describing the effect of \emph{move}
may be stated as:
\begin{align}
  move(robot, Y)~~\mathbf{causes}~~\{loc(robot, Z): neighbor(Z,
  Y)\}% \nonumber\\ \mathbf{if}~~loc(robot, X)
\end{align}
where $Y, Z$ are cells. This causal law states that moving to a cell
can cause the robot to be in one of the neighboring
cells\footnote{This is a special case of a non-deterministic causal
  law defined in extensions of AL with non-boolean fluents, i.e.,
  functions whose values can be elements of arbitrary finite
  domains.}.
% Let $f$ be such a function and $R$ be a subset of possible values of
% $f$. A non-deterministic causal law: $a~~\mathbf{causes}~~f =\{y :
% y\in R\}~~\mathbf{if}~~body$ says that if action $a$ is executed in a
% state satisfying the $body$ then $a$ causes $f$ to take a random value
% satisfying $R$. For instance, if $\{y1,y2\}$ is the subset of values
% satisfying $R$, then: $a~~\mathbf{causes}~~f= y1~\mathbf{or}~f= y2$.
% In the absence of non-boolean fluents, this law is written as this is
% written as: $a~~\mathbf{causes}~~\{f(Y) : R(Y)\}~~\mathbf{if}~~body$.
The LL includes an additional action \emph{search} that enables robots
to search for objects in cells; the corresponding causal laws and
constraints may be written as:
\begin{align}
  &search(cell, object)~~\mathbf{causes}~~searched(cell, object)\\\nonumber
  &found(object, cell)~~\mathbf{if}~~searched(cell, object), \\\nonumber
  &~~~~~~~~~~~~~~~~~~~~~~~~~~~~~~~~~~loc(object, cell)\\\nonumber
  &found(object, room)~~\mathbf{if}~~part\_of(cell, room), \\\nonumber
  &~~~~~~~~~~~~~~~~~~~~~~~~~~~~~~~~~~~~found(object, cell)\\\nonumber
  &continue\_search(room, object)~\mathbf{if}~\lnot found(object,room),\\\nonumber
  &~~~~~~~~~~~~~~~~~~~part\_of(cell, room),\lnot searched(cell, object)
\end{align}
We also introduce a defined fluent \emph{failure} that holds iff the
object under consideration is not in the room that the robot is
searching---this fluent is defined as:
\begin{align}
  \label{eqn:fail}
  &failure(object, room)~~\mathbf{if}~~loc(robot, room), \\\nonumber
  &~~~~\lnot continue\_search(room, object), \lnot found(object, room)
\end{align}
% In the case of failure, the robot reports: $obs(loc(object, room),
% false)$ to the HL history. 
This completes the action theory that describes $\tau_L$. The states
of $\tau_L$ can be viewed as extensions of states of $\tau_H$ by
physically possible fluents and statics defined in the language of LL.
Moreover, for every HL state-action-state transition $\langle \sigma,
a, \sigma^\prime\rangle$ and every LL state $s$ compatible with
$\sigma$ (i.e., $\sigma \subset s$), there is a path in the LL from
$s$ to some state compatible with $\sigma^\prime$.

Unlike the HL system description in which effects of actions and
results of observations are always accurate, the action effects and
observations in the LL are only known with some degree of probability.
The state transition function $T: S\times A\times S' \to [0, 1]$
defines the probabilities of state transitions in the LL. % The bottom
% part of Figure~\ref{fig:state-transition} shows some LL state
% transitions. 
Due to perceptual limitations of the robot, only a subset of the
fluents are observable in the LL; we denote this set of fluents by
$Z$. Observations are elements of $Z$ associated with a probability,
and are obtained by processing sensor inputs using probabilistic
algorithms. The observation function $O: S\times Z \to [0, 1]$ defines
the probability of observing specific observable fluents in specific
states.  Functions $T$ and $O$ are computed using prior knowledge, or
by observing the effects of specific actions in specific states (see
Section~\ref{sec:exp-setup}).

States are partially observable in the LL, and we introduce (and
reason with) \emph{belief states}, probability distributions over the
set of states. Functions $T$ and $O$ describe a probabilistic
transition diagram defined over belief states.  The initial belief
state is represented by $B_0$, and is updated iteratively using
Bayesian inference:
\begin{align}
  \label{eqn:belief-update}
  B_{t+1}(s_{t+1}) \propto O(s_{t+1},o_{t+1})\sum_{s}
  T(s,a_{t+1},s_{t+1})\cdot B_t(s)
\end{align}
The LL system description includes a reward specification $R: S\times
A \times S' \to \Re$ that encodes the relative cost or \emph{value} of
taking specific actions in specific states. % For
% instance, the reward for action at time $t$, $a_t$, may be defined as
% the reduction in entropy between belief state $B_{t-1}$ and the belief
% state $B_{t}$ after executing action $a_t$.
% \begin{align}
%   R(a_{t}) = \sum_{k} b^k_{t} log(b^k_{t}) - \sum_{j} b^j_{t-1}
%   log(b^j_{t-1})
% \end{align}
% For instance, movement and sensing actions may be assigned a negative
% reward based on the time required to execute them, while the reward
% for terminating the LL action sequence depends on whether the robot
% has achieved the desired goal. 
Planning in the LL then involves computing a \emph{policy} that
maximizes the reward over a planning horizon. This policy maps belief
states to actions: $\pi: B_t\mapsto a_{t+1}$. We use a point-based
approximate algorithm to compute this policy~\cite{Ong:ijrr2010}.  In
our illustrative example, an LL policy computed for HL action
\emph{move} is guaranteed to succeed, and that the LL policy computed
for HL action \emph{grasp} considers three LL actions: \emph{move},
\emph{search}, and \emph{grasp}. Plan execution in the LL corresponds
to using the computed policy to repeatedly choose an action in the
current belief state, and updating the belief state after executing
that action and receiving an observation. We henceforth refer to this
algorithm as ``POMDP-1''.

Unlike the HL, history in the LL representation consists of
observations and actions over one time step; the current belief state
is assumed to be the result of all information obtained in previous
time steps (first-order Markov assumption). In this paper, the LL
domain representation is translated automatically into POMDP models,
i.e., specific data structures for representing the components of
$\mathcal{D}_L$ (described above) such that existing POMDP solvers can
be used to obtain action policies.

We observe that the coupling between the LL and the HL has some key
consequences. First, for any HL action, the relevant LL variables are
identified automatically, improving the computational efficiency of
computing the LL policies.  Second, if LL actions cause different
fluents, these fluents are independent.  Finally, although defined
fluents are crucial in determining what needs to be communicated
between the levels of the architecture, they themselves need not be
communicated.

%%%%%-----------------------------------------------------------------------------
\subsection{Control loop}
\label{sec:arch-loop}

%\incmargin{1em}
\begin{algorithm}[tcb]
  \caption{Control loop of architecture}
  \label{alg:hierPlan}
  \DontPrintSemicolon

  \Indm
  \KwIn{The HL and LL domain representations, and the specific task
    for robot to perform.}
  \Indp
  \BlankLine

  LL observations reported to HL history; HL initial state
  ($s^{H}_{init}$) communicated to LL.\;

  Assign goal state $s^{H}_{goal}$ based on task.\;
  
  Generate HL plan(s).\label{alg:replan}\;

  \If{multiple HL plans exist} {
    Send plans to the LL, select plan with lowest (expected) action
    cost and communicate to the HL.\;
  }

  \If{HL plan exists} {
    
    \For{$a_i^H \in$ HL plan: $i \in [1, n]$} {

      Pass $a_i^H$ and relevant fluents to LL.\;

      Determine initial belief state over the relevant LL state space.\;

      Generate LL action policy.\;

      \While{$a_i^H$ not completed \textbf{and} $a_i^H$ achievable}{
  
        Execute an action based on LL action policy.\;
    
        Make an LL observation and update belief state.\;
      }
  
      LL observations and action outcomes add statements to HL
      history.\;

      \If{results unexpected} {
        Perform diagnostics in HL.\;
      }

      \If{HL plan invalid} {
        Replan in the HL (line~\ref{alg:replan}). \;
      }

    }
  }
\end{algorithm}
Algorithm~\ref{alg:hierPlan} describes the architecture's control
loop\footnote{We leave the proof of the correctness of this algorithm
  as future work.}. First, the LL observations obtained in the current
location add statements to the HL history, and the HL initial state
($s^{H}_{init}$) is communicated to the LL (line 1). The assigned task
determines the HL goal state ($s^{H}_{goal}$) for planning (line 2).
Planning in the HL provides a sequence of actions with deterministic
effects (line 3).

In some situations, planning in the HL may provide multiple plans,
e.g., when the object that is to be grasped can be in one of multiple
locations, tentative plans may be generated for the different
hypotheses regarding the object's location. In such situations, all
the HL plans are communicated to the LL and compared based on their
costs, e.g., the expected time to execute the plans.  The plan with
the least expected cost is communicated to the HL (lines 4-6). 

If an HL plan exists, actions are communicated one at a time to the LL
along with the relevant fluents (line 9). For HL action $a_i^H$, the
communicated fluents are used to automatically identify the relevant
LL variables and set the initial belief state, e.g., a uniform
distribution (line 10). An LL action policy is computed (line 11) and
used to execute actions and update the belief state until $a_i^H$ is
achieved or inferred to be unachievable (lines 12-15). The outcome of
executing the LL policy, and the LL observations, add to the HL
history (line 16). For instance, if defined fluent \emph{failure} is
true for object $ob_1$ and room $rm_1$, the robot reports:
$obs(loc(ob_1, rm_1),false)$ to the HL history.  If the results are
unexpected, diagnosis is performed in the HL (lines 17-19); we assume
that the robot is capable of identifying these unexpected outcomes. If
the HL plan is invalid, a new plan is generated (lines 20-22); else,
the next action in the HL plan is executed.

%%%%%-----------------------------------------------------------------------------
%%%%%-----------------------------------------------------------------------------
\section{Experimental setup and results}
\label{sec:exp}
This section describes the experimental setup and results of
evaluating the proposed architecture in indoor domains.

%%%%%-----------------------------------------------------------------------------
\subsection{Experimental setup}
\label{sec:exp-setup}
The architecture was evaluated in simulation and on physical robots.
To provide realistic observations in the simulator, we included object
models that characterize objects using probabilistic functions of
features extracted from images captured by a camera on physical
robots~\cite{Li:icar13}. The simulator also uses action models that
reflect the motion of the robot. Specific instances of objects of
different classes were simulated in a set of rooms. The experimental
setup also included an initial training phase in which the robot
repeatedly executed the different movement actions and applied the
visual input processing algorithms on images with known objects. A
human participant provided some of the ground truth data, e.g., labels
of objects in images. A comparison of the expected and actual outcomes
was used to define the functions that describe the probabilistic
transition diagram ($T$, $O$) in the LL, while the reward
specification is defined by also considering the computational time
required by different visual processing and navigation algorithms.

In each trial of the experimental results summarized below, the
robot's goal is to move specific objects to specific places; the
robot's location, target object, and locations of objects are chosen
randomly in each trial. A sequence of actions extracted from an answer
set obtained by solving the SPARC program of the HL domain
representation provides an HL plan. If a robot (\stt{robot1}) that is
in the \emph{office} is asked to fetch a textbook (\stt{tb1}) from the
\emph{main\_library}, the HL plan consists of the following sequence
of actions:
\begin{align*}
  &move(robot1, main\_library)\\ 
  &grasp(robot1, tb1)\\
  &move(robot1, office)\\
  &putdown(robot1, tb1)
\end{align*}
The LL action policies for each HL action are generated by solving the
appropriate POMDP models using the APPL
solver~\cite{Ong:ijrr2010,Somani:nips13}.  In the LL, the location of
an object is considered to be known with certainty if the belief (of
the object's occurrence) in a grid cell exceeds a threshold ($0.85$).

We experimentally compared our architecture, with the control loop
described in Algorithm~\ref{alg:hierPlan}, henceforth referred to as
``PA'', with two alternatives: (1) POMDP-1 (see
Section~\ref{sec:arch-ll}); and (2) POMDP-2, which revises POMDP-1 by
assigning high probability values to defaults to bias the initial
belief states. These comparisons evaluate two hypotheses: (H1) PA
enables a robot to achieve the assigned goals more reliably and
efficiently than using POMDP-1; (H2) our representation of defaults
improves reliability and efficiency in comparison with not using
default knowledge or assigning high probability values to defaults.

%%%%%-----------------------------------------------------------------------------
\subsection{Experimental Results}
\label{sec:exp-results}
\begin{figure}[tbc]
  \begin{center}\hspace*{-1em}
    \includegraphics[width=1.1\columnwidth]{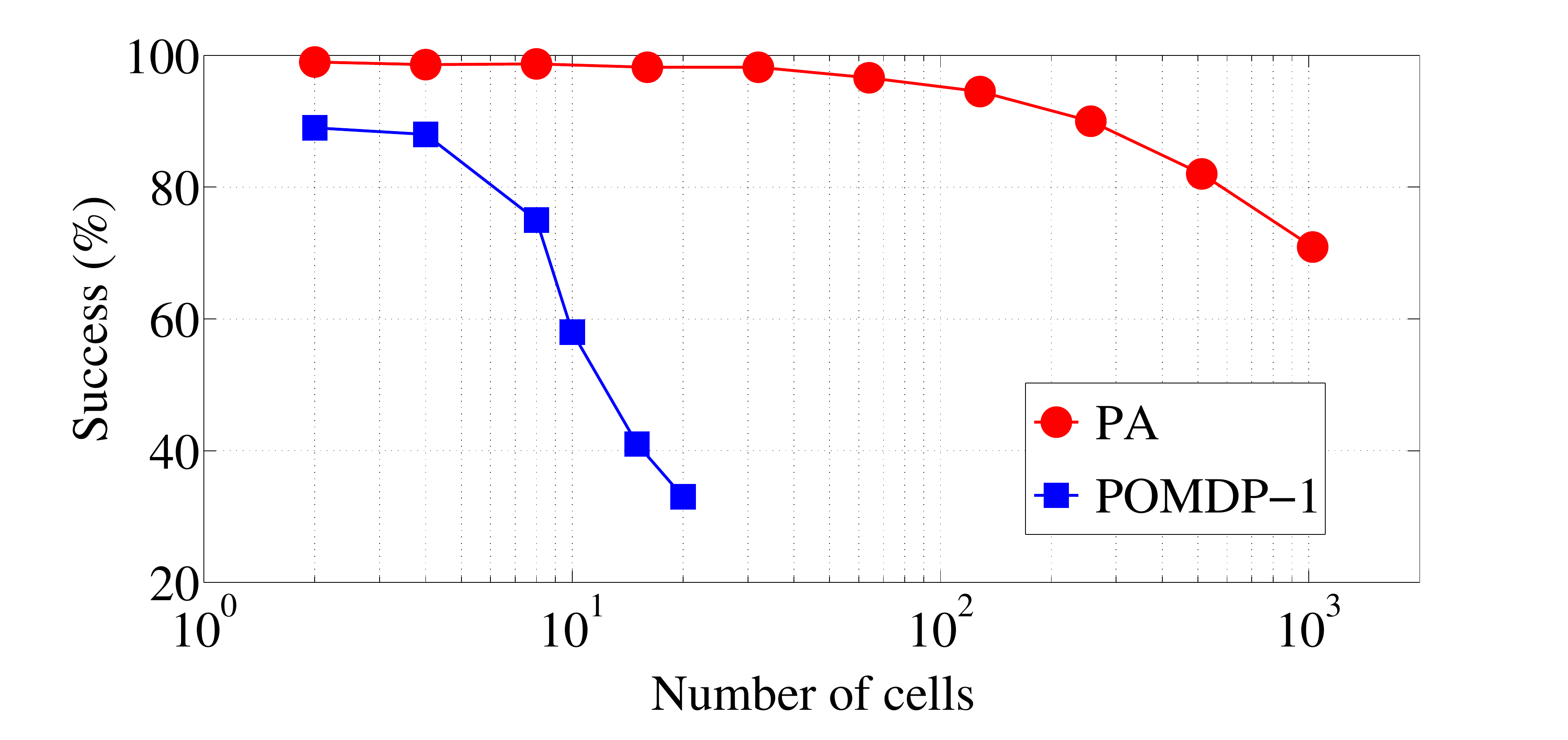}
  \end{center}
  \caption{Ability to successfully achieve the assigned goal, as a
    function of the number of cells in the domain; with a limit on the
    time to compute policies PA significantly increases accuracy in
    comparison with just POMDP-1 as the number of cells in the domain
    increases.}
  \label{fig:accuracy}
\end{figure}

To evaluate H1, we first compared PA with POMDP-1 in a set of trials in
which the robot's initial position is known but the position of the
object to be moved is unknown. The solver used in POMDP-1 is given a
fixed amount of time to compute action policies.
Figure~\ref{fig:accuracy} summarizes the ability to successfully
achieve the assigned goal, as a function of the number of cells in the
domain. Each point in Figure~\ref{fig:accuracy} is the average of
$1000$ trials, and we set (for ease of interpretation) each room to
have four cells. PA significantly improves the robot's ability to
achieve the assigned goal in comparison with POMDP-1. As the number of
cells (i.e., size of the domain) increases, it becomes computationally
difficult to generate good POMDP action policies which, in conjunction
with incorrect observations (e.g., false positive sightings of
objects) significantly impacts the ability to successfully complete
the trials. PA, on the other hand, focuses the robot's attention on
relevant regions of the domain (e.g., specific rooms and cells). As
the size of the domain increases, a large number of plans of similar
cost may still be generated which, in conjunction with incorrect
observations, may affect the robot's ability to successfully complete
the trials---the impact is, however, much less pronounced.

\begin{figure}[tbc]
  \begin{center}\hspace*{-1.7em}
    \includegraphics[width=1.15\columnwidth]{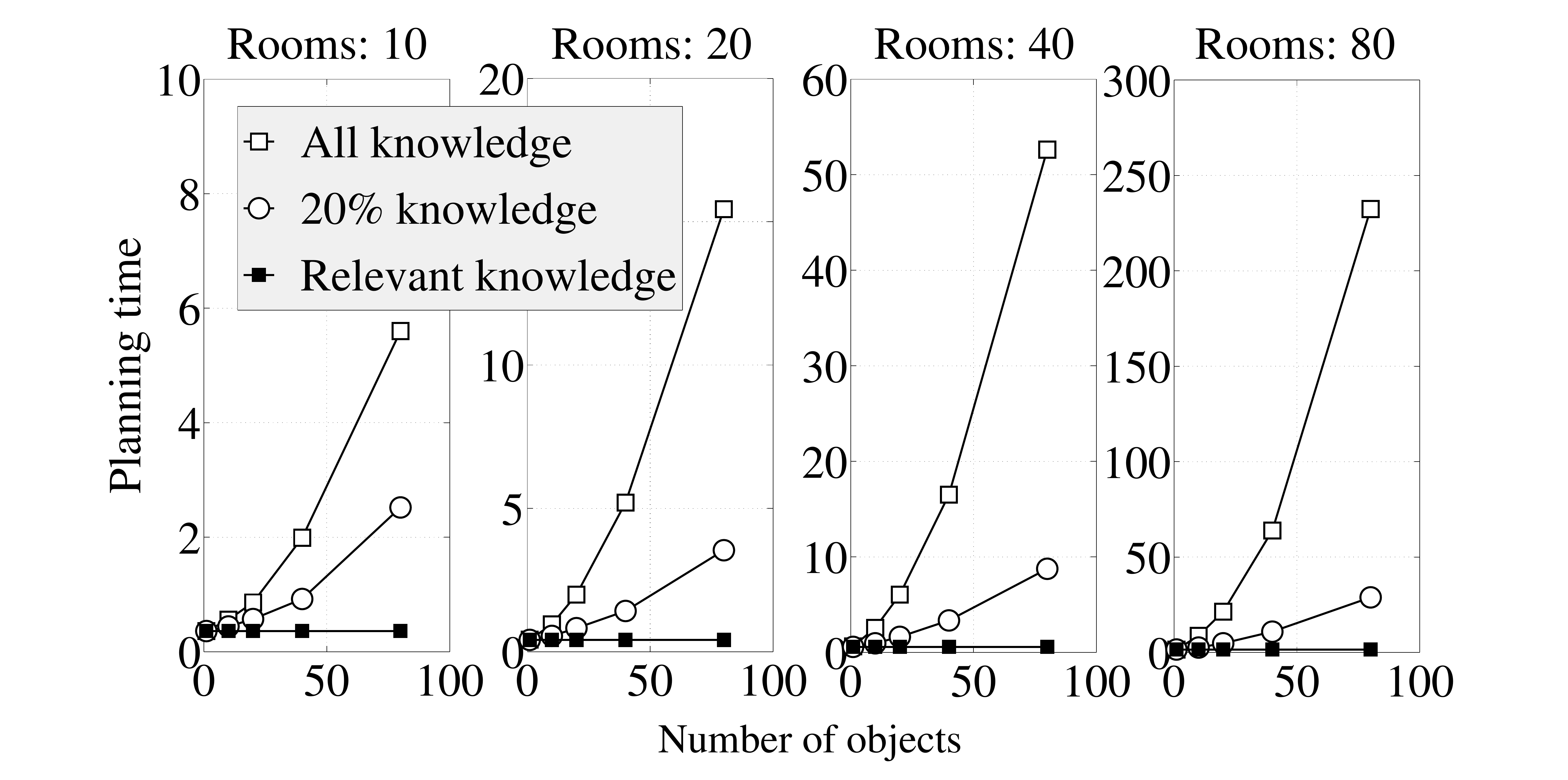}
  \end{center}
  \caption{Planning time as a function of the number of rooms and the
    number of objects in the domain---$PA$ scales to larger number of
    rooms and objects.}
  \label{fig:room-object}
\end{figure}

Next, we computed the time taken by PA to generate a plan as the size
of the domain increases. Domain size is characterized based on the
number of rooms and the number of objects in the domain. We conducted
three sets of experiments in which the robot reasons with: (1) all
available knowledge of domain objects and rooms; (2) only knowledge
relevant to the assigned goal---e.g., if the robot knows an object's
default location, it need not reason about other objects and rooms in
the domain to locate this object; and (3) relevant knowledge and
knowledge of an additional $20\%$ of randomly selected domain objects
and rooms. Figure~\ref{fig:room-object} summarizes these results. We
observe that PA supports the generation of appropriate plans for
domains with a large number of rooms and objects. We also observe that
using only the knowledge relevant to the goal significantly reduces
the planning time---such knowledge can be automatically selected using
the relationships included in the HL system description. Furthermore,
if we only use a probabilistic approach (POMDP-1), it soon becomes
computationally intractable to generate a plan for domains with many
objects and rooms; these results are not shown in
Figure~\ref{fig:room-object}---see~\cite{Sridharan:AIJ10,zhang:TRO13}.

\begin{figure}[tbc]
  \begin{center}\hspace*{-1.5em}
    \includegraphics[width=1.15\columnwidth]{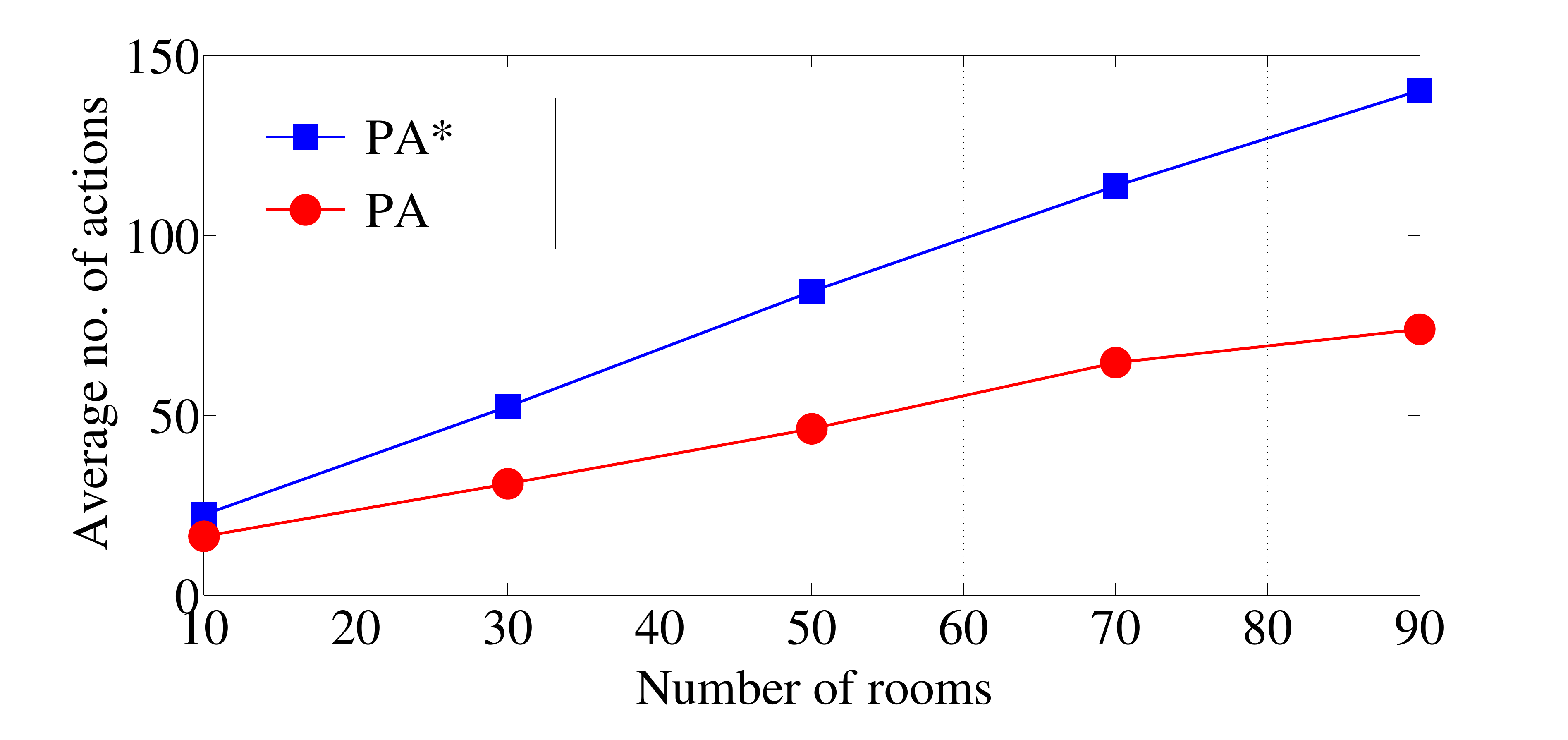}
  \end{center}
  \caption{Effect of using default knowledge---principled
    representation of defaults significantly reduces the number of
    actions (and thus time) for achieving assigned goal.}
  \label{fig:default}
\end{figure}

\begin{figure*}[tbc]
  \begin{center}
    \subfigure[0.6\textwidth][Domain map] {
      \includegraphics[width=0.6\textwidth]{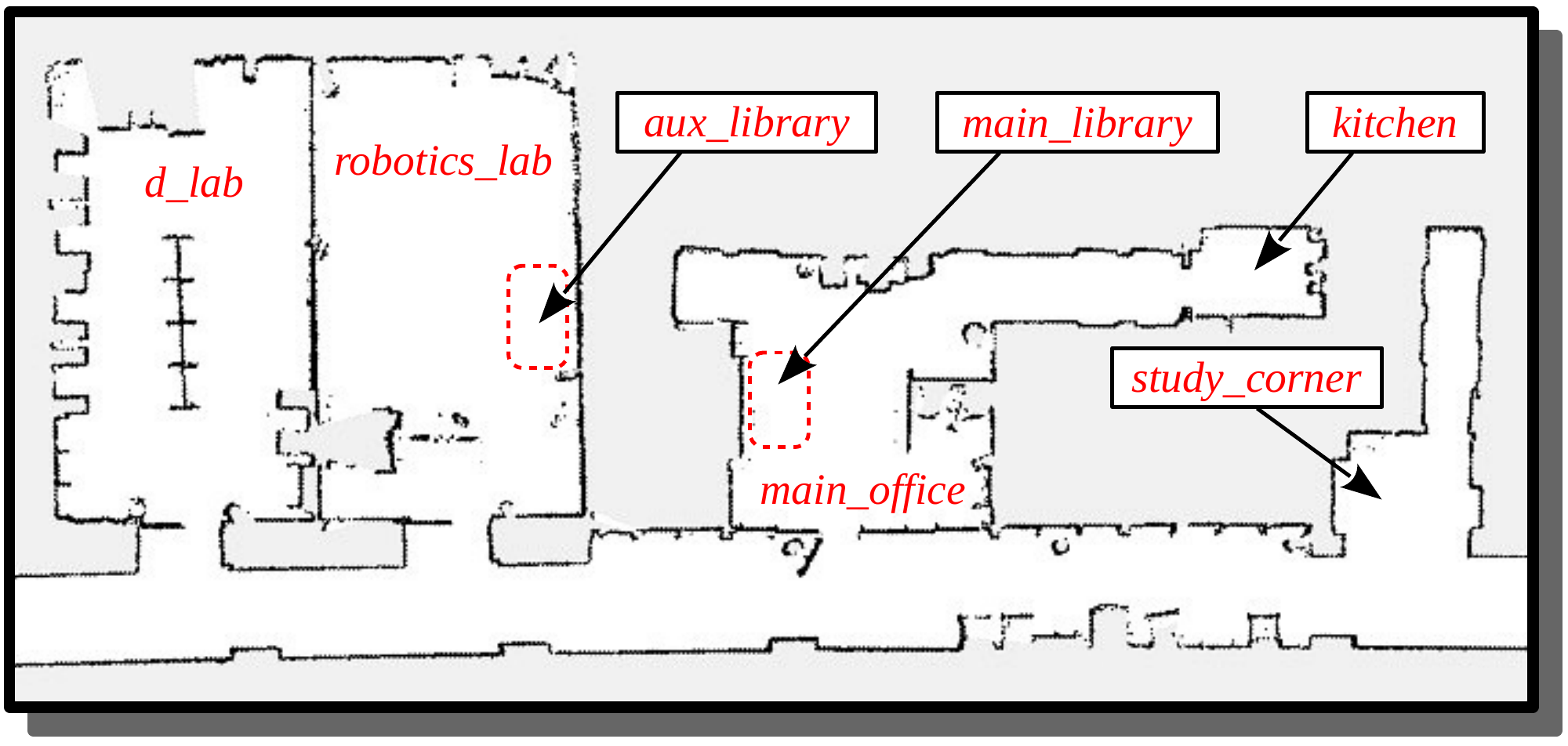}
      \label{fig:map}
    }
    \subfigure[0.3\textwidth][Robot platform] {\hspace*{3em}
      \includegraphics[width=0.11\textwidth]{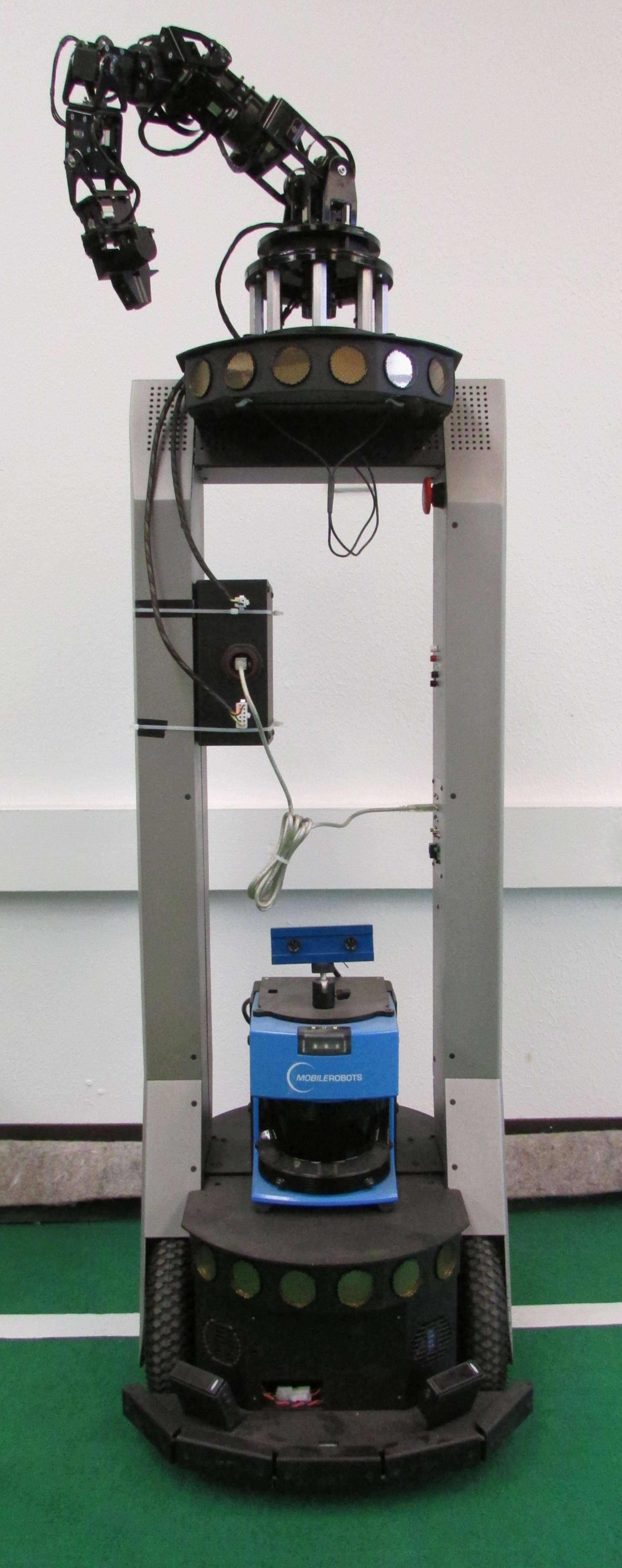} \hspace*{2em}
      \label{fig:robot}
    }
  \end{center}
  \label{fig:map_robot}
  \caption{Subset of the map of the second floor of our department;
    specific places are labeled as shown, and used during planning to
    achieve the assigned goals. The robot platform used in the
    experimental trials is also shown.}
\end{figure*}

To evaluate H2, we first conducted multiple trials in which PA was
compared with $PA^*$, a version that does not include any default
knowledge. Figure~\ref{fig:default} summarizes the average number of
actions executed per trial as a function of the number of rooms in the
domain---each sample point is the average of $10000$ trials. The goal
in each trial is (as before) to move a specific object to a specific
place. We observe that the principled use of default knowledge
significantly reduces the number of actions (and thus time) required
to achieve the assigned goal. Next PA was compared with POMDP-2, which
assigns high probability values to default information and suitably
revises the initial belief state.  We observe that the effect of
assigning a probability value to defaults is arbitrary depending on
multiple factors: (a) the numerical value chosen; and (b) whether the
ground truth matches the default information. For instance, \emph{if a
  large probability is assigned to the default knowledge that books
  are typically in the library, but the book the robot has to move is
  an exception to the default (e.g., a cookbook), it takes a
  significantly large amount of time for POMDP-2 to revise (and
  recover from) the initial belief}. PA, on the other hand, enables
the robot to revise initial defaults and encode exceptions to
defaults.

\paragraph{\underline{Robot Experiments:}}
In addition to the trials in simulated domains, we compared PA with
POMDP-1 on a wheeled robot over $50$ trials conducted on two floors of
our department building. This domain includes places in addition to
those included in our illustrative example, e.g., Figure~\ref{fig:map}
shows a subset of the domain map of the third floor of our department,
and Figure~\ref{fig:robot} shows the wheeled robot platform. Such
domain maps are learned by the robot using laser range finder data,
and revised incrementally over time.  Manipulation by physical robots
is not a focus of this work. Therefore, once the robot is next to the
desired object, it currently asks for the object to be placed in the
extended gripper; future work will include existing probabilistic
algorithms for manipulation in the LL.

For experimental trials on the third floor, we considered $15$ rooms,
which includes faculty offices, research labs, common areas and a
corridor. To make it feasible to use POMDP-1 in such large domains, we
used our prior work on a hierarchical decomposition of POMDPs for
visual sensing and information processing that supports automatic
belief propagation across the levels of the hierarchy and model
generation in each level of the
hierarchy~\cite{Sridharan:AIJ10,zhang:TRO13}. The experiments included
paired trials, e.g., over $15$ trials (each), POMDP-1 takes $1.64$ as
much time as PA (on average) to move specific objects to specific
places. For these paired trials, this $39\%$ reduction in execution
time provided by PA is statistically significant: \emph{p-value} $=
0.0023$ at the $95\%$ significance level.

Consider a trial in which the robot's objective is to bring a specific
textbook to the place named \emph{study\_corner}. The robot uses
default knowledge to create an HL plan that causes the robot to move
to and search for the textbook in the \emph{main\_library}. When the
robot does not find this textbook in the \emph{main\_library} after
searching using a suitable LL policy, replanning in the HL causes the
robot to investigate the \emph{aux\_library}. The robot finds the
desired textbook in the \emph{aux\_library} and moves it to the target
location. A video of such an experimental trial can be viewed
online:\\ \url{http://youtu.be/8zL4R8te6wg}

%%%%%-----------------------------------------------------------------------------
%%%%%-----------------------------------------------------------------------------
\section{Conclusions}
\label{sec:conclusion}
This paper described a knowledge representation and reasoning
architecture for robots that integrates the complementary strengths of
declarative programming and probabilistic graphical models. The system
descriptions of the tightly coupled high-level (HL) and low-level (LL)
domain representations are provided using an action language, and the
HL definition of recorded history is expanded to allow prioritized
defaults.  Tentative plans created in the HL using defaults and
commonsense reasoning are implemented in the LL using probabilistic
algorithms, generating observations that add suitable statements to
the HL history. In the context of robots moving objects to specific
places in indoor domains, experimental results indicate that the
architecture supports knowledge representation, non-monotonic logical
inference and probabilistic planning with qualitative and quantitative
descriptions of knowledge and uncertainty, and scales well as the
domain becomes more complex. Future work will further explore the
relationship between the HL and LL transition diagrams, and
investigate a tighter coupling of declarative logic programming and
probabilistic reasoning for robots.

%%%%%-----------------------------------------------------------------------------
%%%%%-----------------------------------------------------------------------------

\section*{Acknowledgments}
The authors thank Evgenii Balai for making modifications to SPARC to
support some of the experiments reported in this paper. This research
was supported in part by the U.S. Office of Naval Research (ONR)
Science of Autonomy Award N00014-13-1-0766.  Opinions, findings, and
conclusions are those of the authors and do not necessarily reflect
the views of the ONR.

%%%%%-----------------------------------------------------------------------------
%%%%%-----------------------------------------------------------------------------

\small
\bibliographystyle{aaai}
\bibliography{references.bib}

\begin{thebibliography}{}

\bibitem[\protect\citeauthoryear{Balai, Gelfond, and
  Zhang}{2013}]{Balai:lpnmr13}
Balai, E.; Gelfond, M.; and Zhang, Y.
\newblock 2013.
\newblock {Towards Answer Set Programming with Sorts}.
\newblock In {\em {International Conference on Logic Programming and
  Nonmonotonic Reasoning}}.

\bibitem[\protect\citeauthoryear{Balduccini and
  Gelfond}{2003}]{Balduccini:aaaisymp03}
Balduccini, M., and Gelfond, M.
\newblock 2003.
\newblock {Logic Programs with Consistency-Restoring Rules}.
\newblock In {\em Logical Formalization of Commonsense Reasoning, AAAI Spring
  Symposium Series},  9--18.

\bibitem[\protect\citeauthoryear{Baral, Gelfond, and
  Rushton}{2009}]{baral:TPLP09}
Baral, C.; Gelfond, M.; and Rushton, N.
\newblock 2009.
\newblock {Probabilistic Reasoning with Answer Sets}.
\newblock {\em Theory and Practice of Logic Programming} 9(1):57--144.

\bibitem[\protect\citeauthoryear{Baral}{2003}]{Baral:book03}
Baral, C.
\newblock 2003.
\newblock {\em {Knowledge Representation, Reasoning and Declarative Problem
  Solving}}.
\newblock Cambridge University Press.

\bibitem[\protect\citeauthoryear{Chen \bgroup et al\mbox.\egroup
  }{2012}]{Chen:HRI12}
Chen, X.; Xie, J.; Ji, J.; and Sui, Z.
\newblock 2012.
\newblock {Toward Open Knowledge Enabling for Human-Robot Interaction}.
\newblock {\em Journal of Human-Robot Interaction} 1(2):100--117.

\bibitem[\protect\citeauthoryear{Erdem, Aker, and Patoglu}{2012}]{Erdem:ISR12}
Erdem, E.; Aker, E.; and Patoglu, V.
\newblock 2012.
\newblock {Answer Set Programming for Collaborative Housekeeping Robotics:
  Representation, Reasoning, and Execution}.
\newblock {\em Intelligent Service Robotics} 5(4).

\bibitem[\protect\citeauthoryear{Gelfond and Kahl}{2014}]{Gelfond:aibook14}
Gelfond, M., and Kahl, Y.
\newblock 2014.
\newblock {\em {Knowledge Representation, Reasoning and the Design of
  Intelligent Agents}}.
\newblock Cambridge University Press.

\bibitem[\protect\citeauthoryear{Gelfond}{2008}]{Gelfond:book08}
Gelfond, M.
\newblock {2008}.
\newblock {Answer Sets}.
\newblock In {Frank van Harmelen and Vladimir Lifschitz and Bruce Porter}.,
  ed., {\em {Handbook of Knowledge Representation}}. {Elsevier Science}.
\newblock  {285--316}.

\bibitem[\protect\citeauthoryear{Ghallab, Nau, and
  Traverso}{2004}]{Ghallab:plan04}
Ghallab, M.; Nau, D.; and Traverso, P.
\newblock 2004.
\newblock {\em {Automated Planning: Theory and Practice}}.
\newblock San Francisco, USA: Morgan Kaufmann.

\bibitem[\protect\citeauthoryear{Halpern}{2003}]{halpern:book03}
Halpern, J.
\newblock 2003.
\newblock {\em {Reasoning about Uncertainty}}.
\newblock MIT Press.

\bibitem[\protect\citeauthoryear{Hanheide \bgroup et al\mbox.\egroup
  }{2011}]{Hanheide:ijcai11}
Hanheide, M.; Gretton, C.; Dearden, R.; Hawes, N.; Wyatt, J.; Pronobis, A.;
  Aydemir, A.; Gobelbecker, M.; and Zender, H.
\newblock 2011.
\newblock {Exploiting Probabilistic Knowledge under Uncertain Sensing for
  Efficient Robot Behaviour}.
\newblock In {\em International Joint Conference on Artificial Intelligence}.

\bibitem[\protect\citeauthoryear{Hoey \bgroup et al\mbox.\egroup
  }{2010}]{Hoey:CVIU10}
Hoey, J.; Poupart, P.; Bertoldi, A.; Craig, T.; Boutilier, C.; and Mihailidis,
  A.
\newblock 2010.
\newblock Automated {H}andwashing {A}ssistance for {P}ersons with {D}ementia
  using {V}ideo and a {P}artially {O}bservable {M}arkov {D}ecision {P}rocess.
\newblock {\em Computer Vision and Image Understanding} 114(5):503--519.

\bibitem[\protect\citeauthoryear{Kaelbling and
  Lozano-Perez}{2013}]{Kaelbling:IJRR13}
Kaelbling, L., and Lozano-Perez, T.
\newblock 2013.
\newblock {Integrated Task and Motion Planning in Belief Space}.
\newblock {\em International Journal of Robotics Research} 32(9-10).

\bibitem[\protect\citeauthoryear{Laird, Newell, and
  Rosenbloom}{1987}]{Laird:AI87}
Laird, J.~E.; Newell, A.; and Rosenbloom, P.
\newblock 1987.
\newblock {SOAR: An Architecture for General Intelligence}.
\newblock {\em Artificial Intelligence} 33(3).

\bibitem[\protect\citeauthoryear{Langley and Choi}{2006}]{Langley:aaai06}
Langley, P., and Choi, D.
\newblock 2006.
\newblock {An Unified Cognitive Architecture for Physical Agents}.
\newblock In {\em {The Twenty-first National Conference on Artificial
  Intelligence (AAAI)}}.

\bibitem[\protect\citeauthoryear{Leone \bgroup et al\mbox.\egroup
  }{2006}]{Leone:TOCL06}
Leone, N.; Pfeifer, G.; Faber, W.; Eiter, T.; Gottlob, G.; Perri, S.; and
  Scarcello, F.
\newblock 2006.
\newblock {The DLV System for Knowledge Representation and Reasoning}.
\newblock {\em ACM Transactions on Computational Logic} 7(3):499--562.

\bibitem[\protect\citeauthoryear{Li and Sridharan}{2013}]{Li:icar13}
Li, X., and Sridharan, M.
\newblock 2013.
\newblock {Move and the Robot will Learn: Vision-based Autonomous Learning of
  Object Models}.
\newblock In {\em International Conference on Advanced Robotics}.

\bibitem[\protect\citeauthoryear{Milch \bgroup et al\mbox.\egroup
  }{2006}]{milch:bookchap07}
Milch, B.; Marthi, B.; Russell, S.; Sontag, D.; Ong, D.~L.; and Kolobov, A.
\newblock {2006}.
\newblock {BLOG: Probabilistic Models with Unknown Objects}.
\newblock In {\em {Statistical Relational Learning}}. {MIT Press}.

\bibitem[\protect\citeauthoryear{Ong \bgroup et al\mbox.\egroup
  }{2010}]{Ong:ijrr2010}
Ong, S.~C.; Png, S.~W.; Hsu, D.; and Lee, W.~S.
\newblock 2010.
\newblock {Planning under Uncertainty for Robotic Tasks with Mixed
  Observability}.
\newblock {\em International Journal of Robotics Research} 29(8):1053--1068.

\bibitem[\protect\citeauthoryear{Richardson and
  Domingos}{2006}]{Richardson:ML06}
Richardson, M., and Domingos, P.
\newblock 2006.
\newblock {Markov Logic Networks}.
\newblock {\em Machine learning} 62(1).

\bibitem[\protect\citeauthoryear{Rosenthal and Veloso}{2012}]{Rosenthal:aaai12}
Rosenthal, S., and Veloso, M.
\newblock 2012.
\newblock {Mobile Robot Planning to Seek Help with Spatially Situated Tasks}.
\newblock In {\em {National Conference on Artificial Intelligence}}.

\bibitem[\protect\citeauthoryear{Sanner and Kersting}{2010}]{Sanner:aaai10}
Sanner, S., and Kersting, K.
\newblock 2010.
\newblock {Symbolic Dynamic Programming for First-order POMDPs}.
\newblock In {\em {National Conference on Artificial Intelligence (AAAI)}}.

\bibitem[\protect\citeauthoryear{Somani \bgroup et al\mbox.\egroup
  }{2013}]{Somani:nips13}
Somani, A.; Ye, N.; Hsu, D.; and Lee, W.~S.
\newblock 2013.
\newblock {DESPOT: Online POMDP Planning with Regularization}.
\newblock In {\em {Advances in Neural Information Processing Systems (NIPS)}}.

\bibitem[\protect\citeauthoryear{Sridharan, Wyatt, and
  Dearden}{2010}]{Sridharan:AIJ10}
Sridharan, M.; Wyatt, J.; and Dearden, R.
\newblock 2010.
\newblock {Planning to See: A Hierarchical Aprroach to Planning Visual Actions
  on a Robot using POMDPs}.
\newblock {\em Artificial Intelligence} 174:704--725.

\bibitem[\protect\citeauthoryear{Talamadupula \bgroup et al\mbox.\egroup
  }{2010}]{Talamadupula:TIST10}
Talamadupula, K.; Benton, J.; Kambhampati, S.; Schermerhorn, P.; and Scheutz,
  M.
\newblock 2010.
\newblock {Planning for Human-Robot Teaming in Open Worlds}.
\newblock {\em ACM Transactions on Intelligent Systems and Technology}
  1(2):14:1--14:24.

\bibitem[\protect\citeauthoryear{Zhang, Sridharan, and
  Bao}{2012}]{zhang:icdl12}
Zhang, S.; Sridharan, M.; and Bao, F.~S.
\newblock 2012.
\newblock {ASP+POMDP: Integrating Non-monotonic Logical Reasoning and
  Probabilistic Planning on Robots}.
\newblock In {\em International Joint Conference on Development and Learning
  and on Epigenetic Robotics}.

\bibitem[\protect\citeauthoryear{Zhang, Sridharan, and
  Washington}{2013}]{zhang:TRO13}
Zhang, S.; Sridharan, M.; and Washington, C.
\newblock 2013.
\newblock {Active Visual Planning for Mobile Robot Teams using Hierarchical
  POMDPs}.
\newblock {\em IEEE Transactions on Robotics} 29(4).

\end{thebibliography}

%%%%%-----------------------------------------------------------------------------
%%%%%-----------------------------------------------------------------------------

\end{document}